\documentclass[10pt, conference, compsocconf]{IEEEtran}

\usepackage[cmex10]{amsmath}
\usepackage{color}
\usepackage{wrapfig}
\usepackage{graphicx}
\usepackage{array,multirow}

\usepackage{algpseudocode}
\usepackage{algorithm}

\usepackage{lipsum}
\usepackage[pscoord]{eso-pic}

\newcommand{\placetextbox}[3]{
	\setbox0=\hbox{#3}
	\AddToShipoutPictureFG*{
		\put(\LenToUnit{#1\paperwidth},\LenToUnit{#2\paperheight}){\vtop{{\null}\makebox[0pt][c]{#3}}}%
	}%
}%

\begin{document}

\placetextbox{0.65}{1}{International Conference on Pattern Analysis and Intelligent Systems (PAIS'15)}
\placetextbox{0.8}{0.985}{October.26-27, 2015 - Tebessa, Algeria}%

\title{Boundary conditions for Shape from Shading}



\author{\IEEEauthorblockN{Lyes Abada}
\IEEEauthorblockA{LRIA Laboratory\\ 
Computer science Department\\
University of sciences and technology\\(USTHB)\\
 Algiers, Algeria\\
Email: labada@usthb.dz}
\and
\IEEEauthorblockN{Saliha Aouat}
\IEEEauthorblockA{LRIA Laboratory\\ 
Computer science Department\\
University of sciences and technology\\(USTHB)\\
 Algiers, Algeria\\
Email: saouat@usthb.dz}
\and
\IEEEauthorblockN{Omar el farouk Bourahla}
\IEEEauthorblockA{LRIA Laboratory\\ 
Computer science Department\\
University of sciences and technology\\(USTHB)\\
 Algiers, Algeria\\
Email: obourahla@ymail.com}}

\maketitle

\begin{abstract}

The Shape From Shading is one of a ​​computer vision field. It studies the 3D reconstruction of an object from a single grayscale image. The difficulty of this field can be expressed in the local ambiguity (convex / concave). J.Shi and Q.Zhu have proposed a method (Global View) to solve the local ambiguity. This method based on the graph theory and the relationship between the singular points. In this work we will show that the use of singular points is not sufficient and requires further information on the object to resolve this ambiguity.

\end{abstract}


\section{Introduction}
The three-dimensional representation of an object has been used in several fields such as bioinformatics, image synthesis and other domains. Several methods have been proposed to reconstruct an object from one image such as the Shape From Shading (SFS) or two or more images such as the stereo-vision or the stereo-photometry. The Shape From Shading is the reconstruction of a 3D object from a single grayscale image. Horn \cite{horn_1} is the first who introduced the notion of SFS in the 70s. The difficulty of solving the SFS resides in the local ambiguity \cite{abada_3,shi_1} (convex / concave), it lead the researchers to impose constraints and limitations on the surface of the 3D object, the light source and camera model, as well as other information on the solution (eg. the singular points \cite{kimmel_1} and Boundary Condition \cite{abada_2,prados_1}).

D.Durou \cite{denis_1} classified the resolution methods of the SFS into three classes:

The first class are the methods of resolution by the partial differential equations methods (PDE), in witch the SFS is represented by the PDE like Eikonal, Hamilton-Jacobi equations. Among these methods the Characteristic Strips Expansion \cite{horn_1}. Bruss \cite{bruss_1} provides the first solution of the Eikonal equation by Power Series Expansion. Kimmel and Buckstein \cite{kimmel_2} propose a global solution using the Level-set method. Abada and Aouat \cite{abada_3} propose a new resolution method using the tabu search. Rouy and Tourin \cite{rouy_1} propose a method that use the theory of the viscosity solutions to solve the PDE of the first order. Prados \cite{prados_1,prados_2,prados_3} shows several scheme for solving Hamilton-jacoubi equation(PDE) in both cases the perspective and the parallel projections. 

The second class are methods which approximate the image irradiance equation. This category is divided into two subclass: methods of local resolutions \cite{abada_2} in which the approximation of the variation of the surface in each point is through the point and its neighbors. The second subclass, are the methods of linear resolution. These methods approximate the reflectance equation (non-linear) by a linear equation. Generally the methods of this category are easy to implement but do not treat the local ambiguity (convex/concave)

The third category concerns the optimization methods. These methods are based on the the variational approaches.

This paper is organized as follow:

In Section 2, we will show the basic notion of the image formation used to solve the shape from shading.

In Section 3, we will give an overview of the Global View method proposed by Shi and Zhu\cite{shi_1}.

In Section 4, we will discuss the problems of the Global View method and we will detail the improvements proposed to solve these problems.

In section 5, we will show the obtained results and we will compare our results with the results obtained by Shi and Zhu.

\section{Image formation}
Equation \ref{eq:eq_1_2} represents the basic equation of the images formation (for more details see \cite{abada_1,abada_2}):
\begin{eqnarray}
E=\frac{\alpha}{4}(\dfrac{p}{f})^{2}I\cos ^{4}\alpha L \label{eq:eq_1_2}
\end{eqnarray}

\begin{figure}
\begin{center}
\includegraphics[scale=0.54]{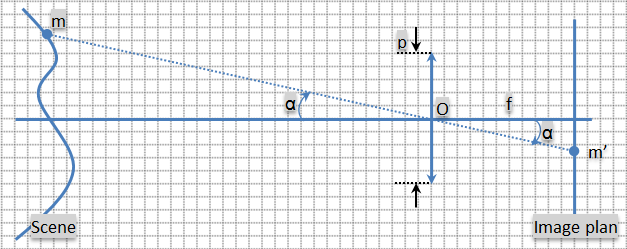}
\caption{Perspective model of the camera}
\end{center}
\end{figure}

The luminance of a Lambert surface can be expressed as follows:

\begin{eqnarray}
L=\dfrac{\rho}{\pi}(\overrightarrow{N}.\overrightarrow{S})\label{eq:eq_5}
\end{eqnarray}

\begin{eqnarray}
N = \dfrac{-p,-q,1}{\sqrt{p^{2}+q^{2}+1}} \label{eq:eq_1111}
\end{eqnarray}
$(p,q)$ is the gradient of $u$.

From (\ref{eq:eq_1_2}),(\ref{eq:eq_5}),(\ref{eq:eq_1111})  we can show that (For more details see \cite{abada_1,abada_2}):

\begin{eqnarray}
E = \dfrac{E_{max}}{\sqrt{p^{2}+q^{2}+1}} \label{eq:eq_eikonal}
\end{eqnarray}

Equation \ref{eq:eq_eikonal} is called "Eikonal equation", we will use it later in this paper.

Several works have proposed to solve the Eikonal equation (eq.\ref{eq:eq_eikonal}). Among these methods the Fast Marching Method(FMM)\cite{rouy_1}. If the minimum distance (shortest path) between two points on the surface is monotony, the difference of depth $D(p_{1},p_{2})$ between theme can be computed. 

$D(p_{1},p_{2})$ can be computed using the following equation \cite{rouy_1}:
\begin{eqnarray}
D(p_{1},p_{2})= inf_{l\in L} \{d(p_{1},p_{2})= \int_l \sqrt{\dfrac{1}{E(s)^{2}}ds} \} \label{eq:eq_1223}
\end{eqnarray}

$p_{1},p_{2}$ are the vertex of the edge extremities.

$D(p_{1},p_{2})$ is the difference in altitude of $p_{1},p_{2}$.

And  L contains all the paths from $p_{1},p_{2}$.

\section{Global View method}

The method Proposed by Shi and Zhu \cite{shi_1} (global view) is very interesting, it is based on the graph theory and the relationships between the singular points. Such as the singular points represent the vertex and the relationship between these points represent the edges of the graph. There is a relationship between two singular points only if the minimum distance (shortest path that check equation \ref{eq:eq_1223}) between two singulars points is monotone (increasing or decreasing). Kimmel  and Bruckstein \cite{shi_1} observed that the singular points and the relationships between them are very important to determine the local ambiguity.  Shi and Zhu model the Shape From Shading problem as a max-cut problem and they use a max-cut solution to solve the SFS problem, this solution is very slow (NP-complete) with big graph (more than ten singular points)  but it solves partially the local ambiguity. The weight of the edge represent the difference in altitude ($\Delta Z$) between the vertex of the two extremities. It can be computed by solving the Eikonal equation (Equation \ref{eq:eq_eikonal}) using the shortest path that check the equation \ref{eq:eq_1223}.

We denote:\\
$G=<V,E,W>$ a graph of n vertex and m edges such that:\\
$V= \{v_{i}/ i=1..n\}$ set of singular points; \\
$E=\lbrace e_{ij}/ i,j \in \lbrace 1..n\rbrace\rbrace:$ set of edges;\\
$W=\lbrace w_{ij}/ i,j \in \lbrace 1..n \rbrace\rbrace:$ set of weight;\\
$D=\lbrace d_{ij} =\pm 1 / i,j \in \lbrace1..n \rbrace\rbrace:$ the graph configuration.\\

The configuration of the edge is the direction of the monotony of the edge (- is down and + is up). The configuration of the graph is the configuration of all the edges of this graph.

\begin{figure}
\begin{center}
\includegraphics[scale=0.6]{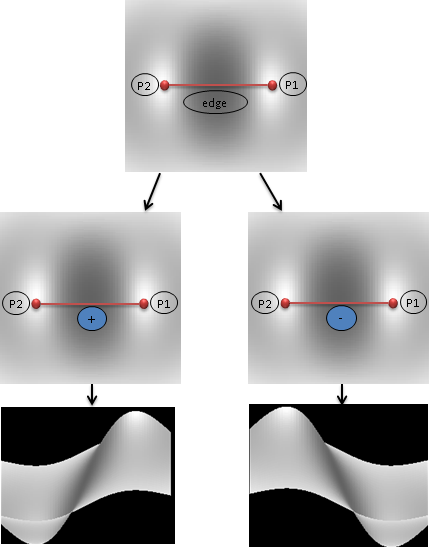}
\caption{local ambiguity for a graph configuration and its reverse }
\label{ambe1}
\end{center}
\end{figure}

A configuration represents a possible solution. Figure (\ref{ambe1}.a) shows the image of the 3D surface generated by the function $f(x,y)=x*exp(-x^{2}-y^{2})$. It contains two singular points (p1,p2), one is a local maximum and the other is a local minimum. If the shortest path between the two points is monotone, Then there exists an edge between the two points. Two possible configurations is observed, the first is the path from p1 to p2 is increasing (the configuration (p1, p2) is "+") or the reverse (the configuration (p1, p2) is "-"). Both configurations generates two different objects but both form the same original picture. So we cannot determine the ambiguity using only the singular points we need an additional information. 

\begin{figure}
\begin{center}
\includegraphics[scale=0.6]{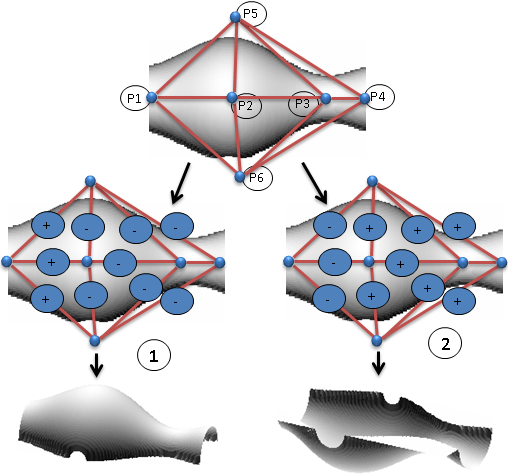}
\caption{local ambiguity for a graph with multiple edge}
\label{ambe2}
\end{center}
\end{figure}

We analyze a more complex example, Figure (\ref{ambe2}) shows the graph generated from the image of the silt, it shows also a possible configuration of the graph of the silt. Note that the reverse configuration is also a solution of the graph but it generates a reverse 3D object. There are other images that can generate multiple configurations that construct several different objects, but these objects form the same original image, and one object that is the correct.

\section{The proposed solution:}

From the above examples we can see that whenever the number of edges between the singulars points increases, the ambiguity decreases.

\begin{figure}
 \begin{center}
 \begin{tabular}{c}
 \includegraphics[height=0.17\textheight]{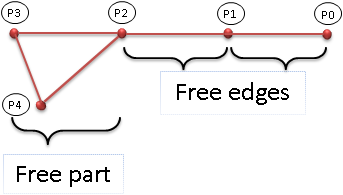}\\
 a\\
  \includegraphics[height=0.17\textheight]{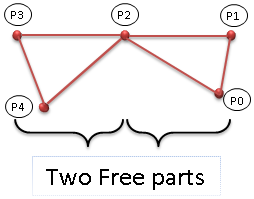}\\
 b
 \end{tabular}
 \end{center}
 \caption{Free parts and Free edges notions}
\label{fig:free}
\end{figure}

We can distinguish three types of ambiguity: 

- The edge Ambiguity: 
If two singular points are connected by a single edge (there is no other path from the first point to the second) then the two possible configurations (+/-) of the edge can be resolved by the "max-cut" problem  for this edge (Figure \ref{fig:free}.a) (we call this edge in this work "Free edge").

- The parts Ambiguity :
If two parts of the graph are connected by one and only one path or vertex (Figures \ref{fig:free}.a,\ref{fig:free}.b), then the configurations of both parts are independent . (We call these parts of the graph "Free part"). A free part may have two solutions the best configuration (computed by max-cut) and its reverse.

- The Global Ambiguity:
In the best case, where the graph consists of a single free part and there is no free edge in this case there is only two solutions the configuration computed by max-cut and its inverse.

Note:
Free edge is a particular case of the Free part, such that this part consists from only two singular points.
If all singular points of the graph are connected by at least two paths then the Global graph composed by only one free part.

\cite{abada_2,prados_1} show the requirement of knowledge of additional information for the uniqueness of the solution, this information represents the Boundary Condition (BC). The BC represents the information of all the pixels around the 3D object. Two types of BC are much used in the SFS field: The Dirichlet BC, the information used in this type is the depth of each outline pixel, and the Neumann BC type such as the information used is the normal vector of the surface or depth gradient.

To determine the ambiguity of using the singular points, we do not need to use all points of the BC. It is sufficient to know which of the two configurations are possible for each free edge or free part of the graph is correct, instead of knowing of all information of the BC. 

\begin{figure}
\begin{center}
\includegraphics[scale=0.7]{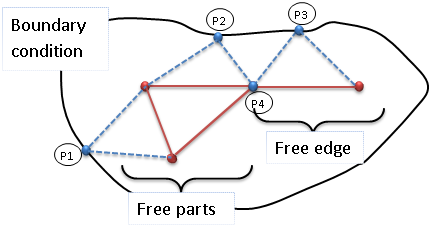}
\caption{Using additional points to determine the ambiguity}
\label{add}
\end{center}
\end{figure}

To solve the problem of ambiguity, 

we propose the knowledge for each free part or free edge two points with a different depth on the BC. We add for each point of BC an edge to the extremities of the free edge. We choose an arc for each free part and we connect it with a BC point by an edge. We can use the same two points for several free edges or free parts. We can also use only one point on the BC if there is a free edge or free part neighbor already computed (see figure \ref{add}). 

\begin{figure}
 \begin{center}
 \begin{tabular}{c}
 \includegraphics[height=0.1\textheight]{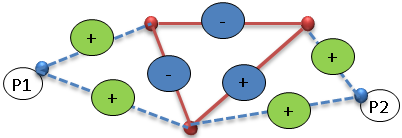}\\
 a\\
  \includegraphics[height=0.1\textheight]{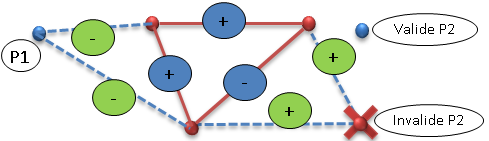}\\
 b
 \end{tabular}
 \end{center}
 \caption{The configuration and the reverse configuration of a graph}
\label{fig:proof}
\end{figure}

proof:
On fixing one point of the two used on the BC. We choose one of the two best possible configurations and we compute the second point. If the configuration is not correct, the second point is shifted with respect to their correct position, in this case we take the reverse configuration (see figure \ref{fig:proof}).

\section{Experiments}

Figures \ref{ambe1} and \ref{ambe2} show two examples for the Global ambiguity. In this section we will show results for the application of the "global view" method on some ambiguous images, and the obtained results when we add the information of the BC. 

\begin{figure}
 \begin{center}
 \begin{tabular}{cc}
 \includegraphics[height=0.15\textheight]{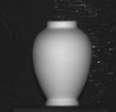}&
  \includegraphics[height=0.18\textheight]{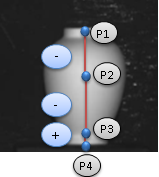}\\
  a&b\\
 \end{tabular}
 \end{center}
 \caption{Application of the max-cut to the image of silt}
\label{fig:i1_m}
\end{figure}

\begin{figure}
 \begin{center}
  \includegraphics[height=0.2\textheight]{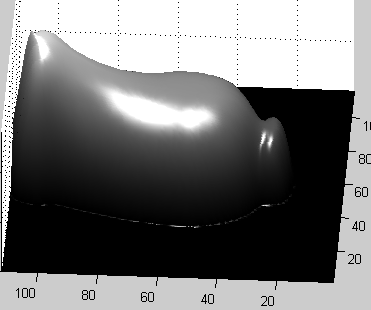}\\
 \end{center}
 \caption{Result of application of the max-cut to the image of silt}
\label{fig:i1_m2}
\end{figure}

\begin{figure}
 \begin{center}
 \begin{tabular}{cc}
 \includegraphics[height=0.15\textheight]{i1.png}&
   \includegraphics[height=0.18\textheight]{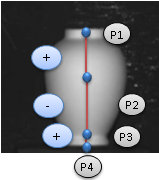}\\
   a&b\\
 \end{tabular}
 \end{center}
 \caption{Application of the max-cut to the image of silt with additional information}
\label{fig:i1_c}
\end{figure}

\begin{figure}
 \begin{center}
 \includegraphics[height=0.14\textheight]{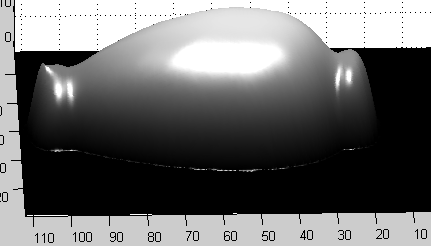}
 \end{center}
 \caption{Result of application of the max-cut to the image of silt with additional information}
\label{fig:i1_c2}
\end{figure}

Figure \ref{fig:i1_m}.a shows the test image of a silt. Figure \ref{fig:i1_m}.b shows the four singular points and the simple graph generated for the silt and the configuration determined by the max-cut method. The 3D object generated by the max-cut configuration of Fig \ref{fig:i1_m}.b have shown in Fig \ref{fig:i1_m2}. In reality the depth of the arc (p1, p2) is Increase but max-cut generates the reverse. This solution is theoretically correct, because the object generated by this configuration form the same image of the silt. But it is false compared with our knowledge. Figure \ref{fig:i1_c}.b shows the configuration obtained using the max-cut and the additional information of the BC. The 3D object generated by the max-cut with additional information have shown in Fig \ref{fig:i1_c2}, it shows that the generated object represents the real silt.

\begin{figure}
 \begin{center}
 \begin{tabular}{cc}
 \includegraphics[height=0.15\textheight]{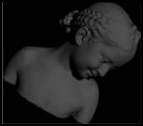}&
   \includegraphics[height=0.15\textheight]{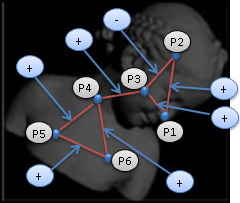}\\
   a&b\\
 \end{tabular}
 \end{center}
 \caption{Application of the max-cut to the image of Figure a}
\label{fig:i2_m}
\end{figure}

\begin{figure}
 \begin{center}
 \includegraphics[height=0.2\textheight]{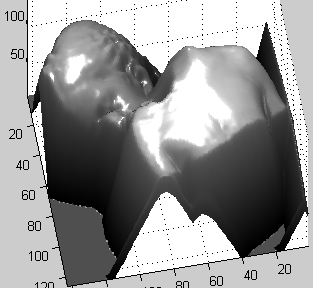}\\
 \end{center}
 \caption{Result of application of the max-cut to the image of Figure \ref{fig:i2_m}.a}
\label{fig:i2_m2}
\end{figure}

\begin{figure}
 \begin{center}
 \begin{tabular}{cc}
 \includegraphics[height=0.15\textheight]{i2.png}&
   \includegraphics[height=0.15\textheight]{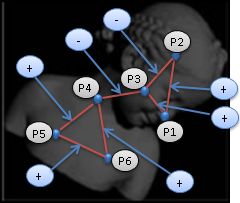}\\
   a&b
 \end{tabular}
 \end{center}
 \caption{Application of the max-cut to the image of Figure a with additional information}
\label{fig:i2_c}
\end{figure}

\begin{figure}
 \begin{center}
 \includegraphics[height=0.2\textheight]{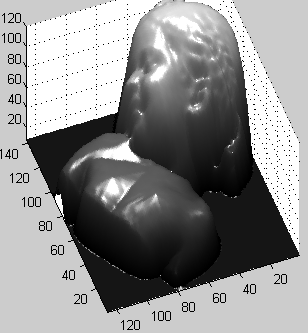}\\
 \end{center}
 \caption{Result of application of the max-cut to the image of Figure \ref{fig:i2_c}.a with additional information}
\label{fig:i2_c2}
\end{figure}

Figure \ref{fig:i2_m}.b shows the graph generated from the image of Figure \ref{fig:i2_m}.a, and the configuration computed by max-cut method. This graph composed by two free parts and one free arc. Max-cut generates the correct configuration for the free part but the free arc is not correct. Figure \ref{fig:i2_m2} shows the 3D object constructed from this configuration. Figure \ref{fig:i2_c}.b shows the configuration generated by the max-cut with the additional information of the BC. Results shown in Figure \ref{fig:i2_c2} is more correct.

\begin{figure}
 \begin{center}
 \begin{tabular}{cc}
 \includegraphics[height=0.2\textheight]{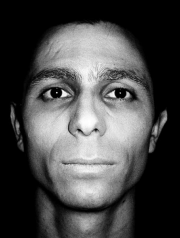}&
    \includegraphics[height=0.2\textheight]{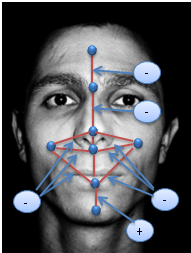}\\
    a&b
 \end{tabular}
 \end{center}
 \caption{Application of the max-cut to the facial image \cite{prados_3}}
\label{fig:i3_m}
\end{figure}

\begin{figure}
 \begin{center}
 \includegraphics[height=0.14\textheight]{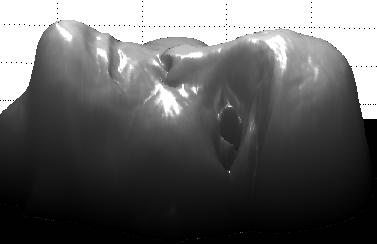}

 \end{center}
 \caption{Result of application of the max-cut to the facial image}
\label{fig:i3_m2}
\end{figure}

\begin{figure}
 \begin{center}
 \begin{tabular}{cc}
 \includegraphics[height=0.2\textheight]{i4.png}&
    \includegraphics[height=0.2\textheight]{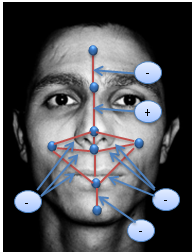}\\
    a&b
 \end{tabular}
 \end{center}
 \caption{Application of the max-cut to the facial image with additional information}
\label{fig:i3_c}
\end{figure}

\begin{figure}
 \begin{center}
 \includegraphics[height=0.14\textheight]{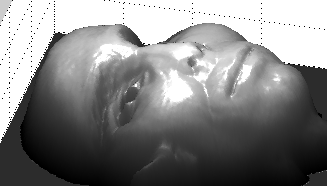}\\
 \end{center}
 \caption{Result of application of the max-cut to the facial image with additional information}
\label{fig:i3_c2}
\end{figure}

Figure \ref{fig:i3_m}.b shows the graph generated from the facial image of figure \ref{fig:i3_m}.a. It is composed from one free part and three free arcs. the configuration computed by the max-cut forms the 3D object of Figure \ref{fig:i3_m2}. The more correct configuration is generated using the max-cut and the additional points of the BC is shown in Figure \ref{fig:i3_c}.b. It generates the 3D object of Figure \ref{fig:i3_c2}.

\section{Conclusion}

The Shape From Shading problem is known by its local ambiguity, the use of the singular points can solve this ambiguity, but only in the case where the singular points are interconnecting. In this work we showed that the singular points are necessary but not sufficient to resolve this ambiguity. Contrary to the methods that uses all information about the Boundary Condition we have shown the minimum information necessary to solve the problem. In the future work we will propose a method for detecting the singular points necessary to resolve the local ambiguity.

\end{document}